\documentclass[letterpaper, 10 pt, conference]{ieeeconf}  % Comment this line out
\IEEEoverridecommandlockouts                              % This command is only
\usepackage{comment}
\usepackage{mathbbol}
\usepackage{tabularx}
\usepackage{algorithm}
\usepackage{xcolor, soul}
\sethlcolor{green}

\let\labelindent\relax

\usepackage{etoolbox}
\makeatletter
\patchcmd{\@makecaption}
  {\scshape}
  {}
  {}
  {}
\makeatother
\usepackage[caption=false]{subfig}

\usepackage{algorithmic}
\usepackage{hyperref}
\usepackage{euscript}[mathcal]
\usepackage[inline]{enumitem}
\usepackage{multirow}
\usepackage{amsmath}
\usepackage{esvect}
\usepackage{color}

\usepackage{url}
\usepackage{xcolor}
\hypersetup{
    colorlinks=true,
    linkcolor=blue,
    filecolor=magenta,      
    urlcolor=cyan,
}

\definecolor{newcolor}{rgb}{0.858, 0.188, 0.478}
\usepackage[english]{babel}
\usepackage[caption=false]{subfig}
\usepackage{multirow}
\usepackage{wrapfig}
\usepackage{booktabs}
\usepackage{array}

\usepackage{stfloats}
\usepackage[utf8]{inputenc}
\usepackage[T1]{fontenc}
\usepackage{amsmath}
\usepackage{cite}
\input epsf
\usepackage{graphicx}
\usepackage{adjustbox}
\graphicspath{{./images/}}
\usepackage{amssymb}
\usepackage{longtable}

\usepackage{caption} 

\usepackage{hyperref}
\usepackage{cleveref}

\makeatletter 
\@namedef{ver@float.sty}{3000/12/31}
\makeatother

\usepackage{mathptmx} % assumes new font selection scheme installed
\usepackage{amssymb}  % assumes amsmath package installed

\title{\LARGE \bf
Fine-Tuning Federated Learning-Based Intrusion Detection Systems for Transportation IoT
}

 \author{Robert Akinie$^{*}$, Nana Kankam Gyimah$^{+}$, Mansi Bhavsar$^{*+}$, John Kelly$^{*}$\\
 $^{*}$ North Carolina A\&T State University, Greensboro, North Carolina, US, 27411\\ 
 $^{+}$ South Carolina State University, Orangeburg, South Carolina, US, 29117\\ 
 $^{*+}$ Minnesota State University, Mankato, Minnesota, US, 56001\\ 
 }

\begin{document}

\maketitle
\thispagestyle{empty}
\pagestyle{empty}

%%%%%%%%%%%%%%%%%%%%%%%%%%%%%%%%%%%%%%%%%%%%%%%%%%%%%%%%%%%%%%%%%%%%%%%%%%%%%%%%
\begin{abstract}
The rapid advancement of machine learning (ML) and on-device computing has revolutionized various industries, including transportation, through the development of Connected and Autonomous Vehicles (CAVs) and Intelligent Transportation Systems (ITS). These technologies improve traffic management and vehicle safety, but also introduce significant security and privacy concerns, such as cyberattacks and data breaches. Traditional Intrusion Detection Systems (IDS) are increasingly inadequate in detecting modern threats, leading to the adoption of ML-based IDS solutions. Federated Learning (FL) has emerged as a promising method for enabling the decentralized training of IDS models on distributed edge devices without sharing sensitive data. However, deploying FL-based IDS in CAV networks poses unique challenges, including limited computational and memory resources on edge devices, competing demands from critical applications such as navigation and safety systems, and the need to scale across diverse hardware and connectivity conditions. To address these issues, we propose a hybrid server-edge FL framework that offloads pre-training to a central server while enabling lightweight fine-tuning on edge devices. This approach reduces memory usage by up to 42\%, decreases training times by up to 75\%, and achieves competitive IDS accuracy of up to 99.2\%. Scalability analysis further demonstrates minimal performance degradation as the number of clients increases, highlighting the framework's feasibility for CAV networks and other IoT applications.

\end{abstract}

%%%%%%%%%%%%%%%%%%%%%%%%%%%%%%%%%%%%%%%%%%%%%%%%%%%%%%%%%%%%%%%%%%%%%%%%%%%%%%%%
\section{Introduction}
Advances in Machine Learning (ML), particularly on-device ML in the Internet of Things (IoT), are driving innovations in transportation \cite{ahmad2022data, chang2021survey}. Connected and Autonomous vehicles (CAVs) play a key role in Intelligent Transportation Systems (ITS), utilizing vehicle-to-vehicle (V2V), vehicle-to-infrastructure (V2I), and vehicle-to-everything (V2X) communication to enhance safety and traffic flow \cite{datta2016integrating, eskandarian2019research, machardy2018v2x}. However, these advancements bring critical security challenges, such as cyberattacks and data leakage, necessitating robust Intrusion Detection Systems (IDS) \cite{hataba2022security}. Traditional IDS struggle with evolving threats, increasing interest in ML-based IDS. Federated Learning (FL), a decentralized and privacy-preserving technique, offers a promising solution for resource-constrained CAV networks.

\begin{figure}[h]
    \includegraphics[scale=0.45]{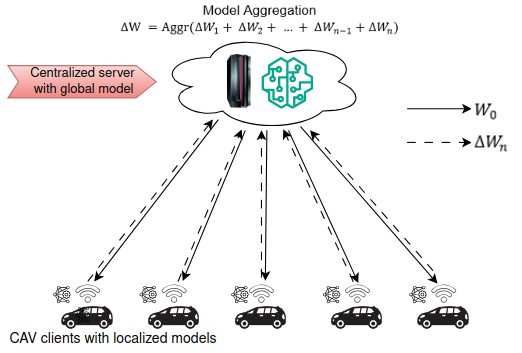}
    \centering
    \caption{Federated Learning Architecture for vehicular networks. CAV clients share model updates (or parameters) from previous training to a centralized server for aggregation into a new update for continuous training. }
    \label{fig:FLarch.png}
\end{figure}

Federated Learning, as illustrated in Figure \ref{fig:FLarch.png}, mitigates the limitations of centralized ML by enabling localized model training without sharing sensitive data \cite{mcmahan2017communication}. This decentralized approach is particularly suitable for distributed systems like IoT networks and CAVs, where it supports real-time adaptability, improved safety, and enhanced intelligence \cite{chellapandi2023federated}. Furthermore, FL’s inherent privacy-preserving nature makes it well-suited for IDS, where safeguarding sensitive data and ensuring real-time threat detection are critical.

The core problem lies in deploying Federated Learning-based Intrusion Detection Systems (FL-based IDS) in resource-constrained environments, particularly in CAV networks, to ensure robust, real-time, and privacy-preserving threat detection. Effectively addressing this problem requires overcoming several unique challenges inherent to these environments:

\begin{enumerate}
    \item \textbf{Resource Constraints:} Limited computational power, memory and energy on CAV edge devices hinder complex on-device model training and inference, reducing the feasibility of FL-based IDS without optimization.
    \item \textbf{Competing Applications:} Concurrent applications like navigation, safety, and infotainment compete for resources, limiting IDS performance in real-time threat detection.
    \item \textbf{Scalability and Network Diversity:} FL frameworks must scale efficiently across diverse CAV networks with varying hardware and connectivity while maintaining consistent IDS performance.
\end{enumerate}

Federated Learning has shown promise for decentralized learning in privacy-sensitive environments like CAVs. McMahan et al. \cite{mcmahan2017communication} demonstrated its potential to preserve data privacy but relied on resource-intensive edge devices, limiting feasibility. Chellapandi et al. \cite{chellapandi2023federated} applied FL to CAVs, highlighting real-time adaptability but overlooking resource constraints. Zawad et al. \cite{zawad2024hyperparameter} addressed these constraints with hyperparameter optimization but failed to explore hybrid server-edge architectures. While showcasing FL’s potential, these studies expose gaps in resource optimization and scalability. This study addresses these gaps by proposing a hybrid server-edge FL framework to reduce edge device burdens while ensuring scalability and adaptability.

To address the challenges of implementing FL-based IDS in resource-constrained environments, this study proposes a \textbf{hybrid server-edge FL framework}. The framework leverages server-side computational power for pre-training while limiting edge device operations to lightweight fine-tuning tasks. This approach optimizes resource utilization while maintaining model adaptability and performance. The primary contributions of this work are as follows:
\begin{enumerate}
    \item \textbf{Hybrid Learning Framework:} We propose a server-edge architecture where the server performs pre-training on proxy data, reducing the computational burden on edge devices. Participating clients fine-tune the model locally on specified model layers using their data, ensuring privacy and adaptability.
    \item \textbf{Improved Efficiency:} Experiments demonstrate that the proposed framework significantly reduces memory overhead and training times on edge devices, achieving optimal model performance of up to a $42\%$ reduction in memory consumption, compared to other frameworks.
    \item \textbf{Scalability and Versatility:} The framework is scalable across diverse CAV networks and IoT domains, accommodating heterogeneous hardware capabilities and varying connectivity conditions while maintaining consistent performance.
\end{enumerate}

The remainder of this paper is structured as follows: Section II discusses related work on FL-based IDS. Section III details the proposed methodology. Section IV presents the datasets, experimental setup, and results. Section V presents the experimental results and discussion. Finally, the conclusion and future work are provided in Section VI.

\section{Related Works}
This section provides an overview of intrusion detection methods for CAVs and Federated Learning, highlighting resource limitations in cross-device FL and the role of Transfer Learning in addressing these challenges.

\subsection{CAV-based IDS and Federated Learning}
Intrusion Detection Systems are critical for addressing the growing security challenges in both external and in-vehicle networks of Connected and Autonomous vehicles. Recent studies have demonstrated the effectiveness of integrating deep learning methods into IDS to detect and mitigate cyber threats \cite{s21144736}. These approaches employ diverse network architectures with varying capabilities \cite{lansky2021deep, aleesa2020review, guha2019one}, offering insights into potential solutions but often neglecting privacy and scalability concerns in distributed environments such as CAVs.

CAV systems face a variety of threats, including network-based attacks (e.g., denial-of-service), physical tampering (e.g., sensor spoofing), software vulnerabilities (e.g., malware injection), data integrity issues (e.g., data tampering or exfiltration), and privacy breaches. These threats create a demand for robust IDS solutions that can handle the complexity and diversity of CAV networks. Federated Learning has emerged as a promising tool for addressing these challenges by enabling collaborative IDS model training without sharing sensitive data \cite{chellapandi2023survey, mcmahan2017communication, olagunju2024privacy}. This decentralized approach enables real-time adaptation to emerging threats, making it particularly suited for CAV environments.

For example, Li et al. \cite{li2020deepfed} proposed DeepFed to detect cyber threats in industrial cyber-physical systems, while Nguyen et al. \cite{nguyen2019diot} developed an anomaly detection-based IDS using FL to autonomously detect compromised IoT devices. Despite these advancements, existing FL-based IDS approaches often overlook resource constraints in edge devices, such as limited memory, computation, and energy. These limitations restrict the scalability and practical deployment of FL-based IDS in real-world CAV environments, necessitating frameworks that optimize resource efficiency.

\subsection{Resource Limitations}
While FL provides privacy-preserving and scalable model training, its implementation is often hampered by resource constraints on edge devices \cite{bonawitz2019towards}. Edge devices typically have limited computational power, memory, energy, and bandwidth,  which must be carefully managed in FL frameworks. Existing FL research addresses these constraints through techniques such as server/client-side hyperparameter tuning for improved inference efficiency \cite{cai2019once, kumar2017resource}, bottleneck reduction \cite{ashraf2020novel}, and model compression \cite{aleesa2020review}. Although effective for individual devices, these methods often fail to consider the collaborative nature of FL, where devices must participate in iterative training cycles.

This gap underscores the need for frameworks that balance resource efficiency with scalability while preserving model accuracy. Addressing these limitations requires innovative approaches that reduce computational and memory overhead without sacrificing the collaborative benefits of FL. Transfer Learning offers a promising direction for achieving this balance, as discussed in the next subsection.

\subsection{Transfer Learning}
Transfer Learning leverages knowledge from a pre-trained model on a source domain and applies it to a target domain. This involves adapting the model through fine-tuning or feature extraction using a smaller dataset relevant to the target task \cite{iman2023review, shin2016deep}. Although feature extraction is memory-efficient, it often sacrifices accuracy by not retaining hidden layer outputs. However, fine-tuning the entire network improves accuracy but increases memory demands, making it unsuitable for resource-constrained devices \cite{cai2020tinytl}.

The aforementioned methods address some resource limitations in Federated Learning, such as memory management and computation efficiency, but they often fail to balance these optimizations with the collaborative and iterative nature of FL training. Furthermore, while individual techniques such as model compression or hyperparameter tuning enhance device-level performance, they do not fully account for the scalability and resource-sharing demands of FL frameworks in real-world distributed environments like CAV networks.

\section{Proposed Methodology}
In this paper, we propose Federated Fine-Tuning (FedFT), a framework that addresses resource constraints in CAV edge computing environments. By leveraging transfer learning, our approach combines server-side pre-training with client-side fine-tuning, significantly reducing computational and memory overhead on edge devices while maintaining accuracy and scalability. This makes the FedFT framework ideal for deploying FL-based IDS in resource-limited settings.

In our framework, the server pre-trains a global intrusion detection model using proxy data and shares the model parameters with participating clients. Clients then perform lightweight fine-tuning only on the classification head using their localized data, freezing the computationally expensive layers. This approach minimizes resource demands while preserving accuracy.

\begin{figure}[h]
    \includegraphics[scale=0.54]{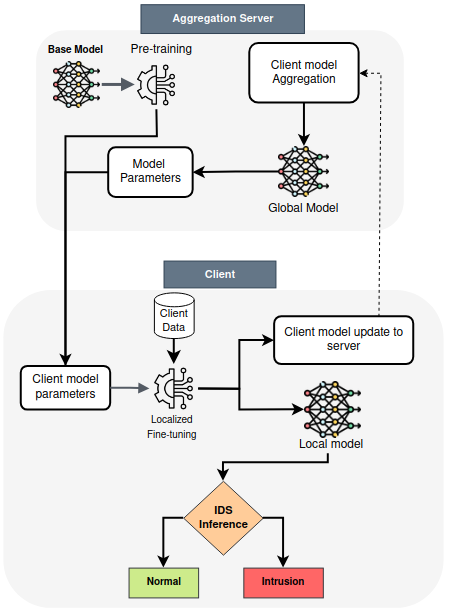}
    \centering
    \caption{The FedFT framework: Server-side pre-training on proxy data, and client-side fine-tuning on localized data for IDS classification and inference.}
    \label{fig:Frame.png}
\end{figure}

As illustrated in Fig.~\ref{fig:Frame.png}, the framework consists of two main components:
\begin{itemize}
    \item A \textbf{central server} that pre-trains the global model, aggregates client updates, and evaluates the federated model after each communication round.
    \item A \textbf{cluster of clients} that receive the pre-trained model, fine-tune the classification head using local data, and send updated parameters back to the server for aggregation.
\end{itemize}

\subsection{Federated Learning Workflow}
We implement the FedFT framework using Flower \cite{beutel2020flower}, a Federated Learning framework that supports large-scale FL experiments in simulated and real-world environments. The workflow of the FedFT system involves the following steps:
\begin{enumerate}
    \item \textbf{Global Model Initialization:} The server pre-trains the global IDS model using proxy data for a specified number of epochs.
    \item \textbf{Model Distribution:} The server distributes the pre-trained model parameters to participating clients.
    \item \textbf{Layer-wise Fine-Tuning on Clients:} Clients freeze the CNN module and fine-tune the lightweight MLP classification head using their respective local datasets.
    \item \textbf{Model Parameter Updates:} Clients send the updated parameters back to the server.
    \item \textbf{Model Aggregation:} The server aggregates the client updates to generate a new global model.
    \item \textbf{Federated Rounds:} The updated global model is redistributed to clients for further localized fine-tuning over multiple communication rounds.
\end{enumerate}

This iterative process ensures that the FedFT framework achieves efficient collaborative training while reducing resource consumption on edge devices. By redistributing the updated global model, the framework also ensures scalability across multiple clients and communication rounds, enabling deployment in diverse and large-scale CAV networks.

\subsection{Model Architecture and Training Process}
\begin{figure}[h]
    \includegraphics[scale=0.28]{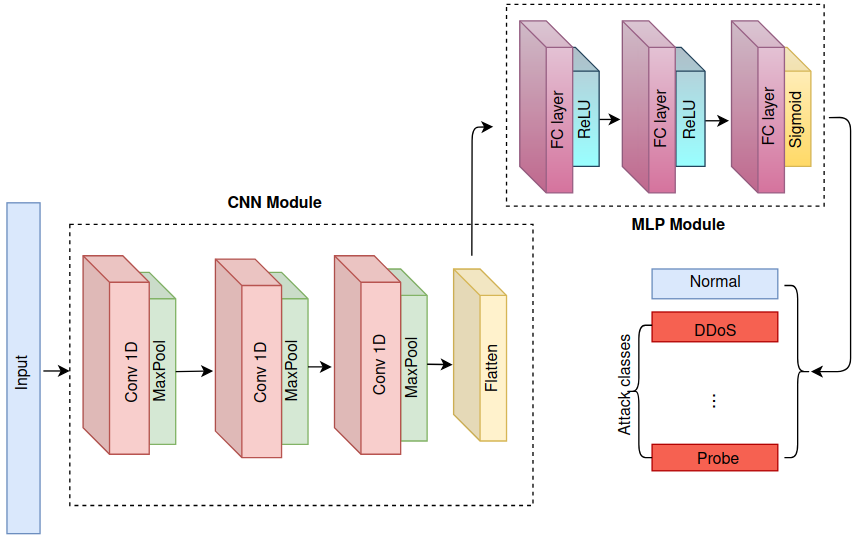}
    \centering
    \caption{IDS model architecture: Pre-trained CNN Module with Conv1D-MaxPooling layers for feature extraction, and a Multilayer Perceptron module with Fully Connected and ReLU layers for fine-tuning.}
    \label{fig:model.png}
\end{figure}

The IDS model architecture combines feature extraction and classification into a lightweight design optimized for transfer learning, as shown in Fig.~\ref{fig:model.png}. The model consists of the following components:
\begin{enumerate}
    \item \textbf{CNN Module:} Three convolutional blocks, each containing a 1-D convolutional layer (Conv1D) for feature extraction and a max-pooling layer for downsampling. The Conv1D layers capture spatial dependencies, while the pooling layers reduce feature dimensions.
    \item \textbf{MLP Module:} Fully connected (FC) layers with Rectified Linear Unit (ReLU) activation functions perform higher-level feature abstraction.
    \item \textbf{Sigmoid Layer:} The final layer applies a sigmoid activation function for binary class classification, normal versus all attack instances grouped under intrusion.
\end{enumerate}

%The model takes as input a feature vector \( x \), representing pre-processed IDS dataset features. The CNN module extracts low-level features, which are flattened and passed to the MLP module for higher-level abstraction. Finally, the sigmoid layer outputs the probability distribution across attack classes, such as Normal, DDoS, and Probe.

\textbf{Training Process:} The training process integrates server-side pre-training with client-side fine-tuning as follows:
\begin{enumerate}
    \item \textbf{Server Initialization:} The global IDS model is pre-trained on the central server using proxy data for a specified number of epochs. The server then shares the model parameters with participating clients via the Flower framework.
    \item \textbf{Client Fine-Tuning:} On each client, the CNN module is frozen, and only the lightweight MLP module is fine-tuned using local data. This significantly reduces computational and memory overhead on resource-constrained edge devices.
    \item \textbf{Parameter Updates and Aggregation:} Clients send the fine-tuned parameters back to the server, where they are aggregated to update the global model.
    \item \textbf{Federated Rounds:} The updated global model is redistributed for further fine-tuning among clients over several communication rounds, ensuring convergence while preserving resource efficiency.
\end{enumerate}

By freezing the CNN module and fine-tuning only the classification head, the FedFT framework minimizes resource usage without compromising detection accuracy. This hybrid approach leverages transfer learning to achieve a balance between resource efficiency, scalability, and model performance, making it well suited for real-world deployment in resource-constrained CAV environments.

\section{Experimental Setting}
This section describes the dataset, the experiment setup, the hyperparameter tuning, and the evaluation metrics of the network intrusion detection model.

\subsection{Dataset}
To evaluate the proposed framework, we use the NSL-KDD dataset \cite{dhanabal2015study}, an improved version of the widely used KDD '99 dataset \cite{tavallaee2009detailed} and a benchmark for intrusion detection systems. Although not explicitly designed for CAV architecture, it includes relevant network attacks, such as Denial of Service (DoS), which are applicable to securing V2X communications in CAV networks.  

The NSL-KDD dataset contains four main categories of attacks:  
\begin{itemize}
    \item \textbf{DoS (Denial of Service):} Disrupts resources to make them unavailable, e.g., SYN flood.  
    \item \textbf{Probe:} Attempts to gather information about network vulnerabilities, e.g., ipsweep and Satan.  
    \item \textbf{R2L (Remote to Local):} Exploits network vulnerabilities to gain unauthorised local access, e.g., Perl and ejection attacks.  
    \item \textbf{U2R (User to Root):} Attempts to escalate privileges to root or superuser, e.g., buffer overflow.  
\end{itemize}

An overview of the dataset distribution is shown in Table \ref{table:nslkdd}. For evaluation, the dataset is split as follows: we utilize the entire testing data as proxy data for pre-training on the server. Among participating devices, we simulate having different data from multiple clients by splitting the original NSL-KDD Train dataset into equal multiple partitions, with each partition representing localized data from each participating client. Pre-processing is then applied to standardize each data partition, followed by feature extraction using principal component analysis (PCA) on each client. We assume that the dataset is independently and identically distributed (IID).

\begin{table}[h!]
\caption{Distribution of attack categories (Normal, DoS, Probe, R2L, U2R) in the NSL-KDD dataset for training (NSL-Train) and testing (NSL-Test) subsets.}

\centering
\begin{tabular}{c c c c c c c}
\hline
         & \textbf{Normal} & \textbf{DoS} & \textbf{Probe} & \textbf{R2L} & \textbf{U2R} & \textbf{Total} \\ \hline
\textbf{NSL-Train} & 67,343          & 45,927        & 11,656         & 995          & 52          & 125,973       \\
\textbf{NSL-Test}  & 9,711           & 7,458         & 2,421          & 2,754         & 200       & 22,543        \\ \bottomrule
\end{tabular}
\label{table:nslkdd}
\end{table}

\subsection{Experiment Setup}
To evaluate the performance of our proposed FedFT framework, we simulate real-world intrusion detection in connected and autonomous vehicle (CAV) ecosystems using edge devices. The experimental setup includes a Flower\cite{beutel2020flower} server, hosted on a Dell Precision Tower, configured with the Federated Averaging\cite{mcmahan2017communication} strategy for aggregating client updates. Flower clients operate on Raspberry Pi 4 devices, using PyTorch as the deep learning framework. Table \ref{table:specification} summarizes the hardware and software specifications.

\begin{table}[h!]
\centering
\caption{Hardware specifications of the devices used in the experimental setup, including the Dell Precision Tower for the Flower server and the Raspberry Pi 4 for client devices.}
\begin{tabular}{c c c}
\hline
\textbf{Device}            & Dell Precision Tower    & Raspberry Pi 4          \\ \hline
\textbf{Memory}            & 48 GB DDR4              & 8 GB LPDDR4 SDRAM          \\ \hline
\textbf{Kernel}            & Linux                    & Linux     \\ 
                            &  (Ubuntu 22.04)         &  (Ubuntu 20.04)              \\ \hline
\textbf{Processor/SoC}     & Intel Xeon W            & Broadcom BCM2711         \\ \hline
\textbf{GPU}               & NVIDIA GeForce        & Broadcom  VideoCore VI   \\ 
                            &   RTX 3050           &           \\ \hline
\end{tabular}
\label{table:specification}
\end{table}

\begin{figure}[h]
    \includegraphics[scale=0.33]{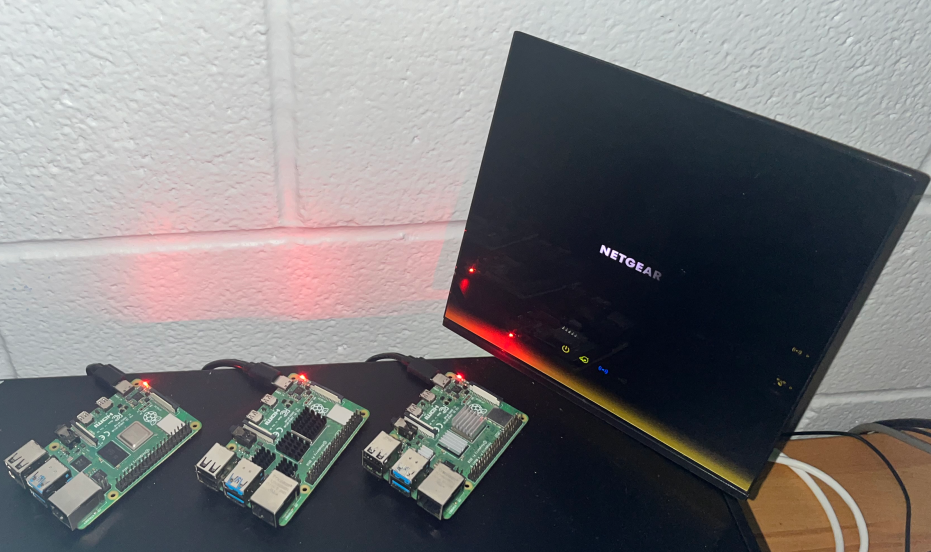}
    \centering
    \caption{Deployment setup showing a Flower server coordinating with three client devices, each representing an independent intrusion detection system (IDS) for training and inference.}
    \label{fig:deployment.png}
\end{figure}

As shown in Fig.~\ref{fig:deployment.png}, the experimental implementation includes three clients, where each client operates as an independent intrusion detection system for CAVs. Motivated by \cite{bhavsar2024fl}, the clients train locally on their respective datasets and perform intrusion detection during inference. Communication between the server and clients occurs over a local area network (LAN) to minimize communication latency.

To comprehensively assess the FedFT framework, we compare it with the following baselines:  
\begin{itemize}
    \item A \textbf{traditional centralized framework}, where all data is collected and processed by a single model.
    \item A \textbf{state-of-the-art FL-IDS framework} \cite{bhavsar2024fl}.
    \item \textbf{Two variations of FedFT}, where different numbers of layers are fine-tuned to analyze their impact on model performance and memory consumption.
\end{itemize}

To ensure reliability, each experiment is repeated three times and the average results are reported across all metrics.

\subsection{Hyperparameter Tuning}
To optimize model performance while balancing resource efficiency, we evaluated several hyperparameters. For client-side fine-tuning, batch sizes of 1, 16, 32, and 64 were tested, with 32 providing the best trade-off between resource consumption and accuracy. Local training was conducted for each FL round with 5, 10, and 15 epochs, with 5 being chosen, as higher values increased computational costs without significant accuracy gains.

Fine-tuning was performed using SGD (Stochastic Gradient Descent) with a learning rate of 0.01 and momentum of 0.9, which were empirically chosen. Communication between the server and clients occurred over a local area network (LAN), and communication latency was not considered in this evaluation.

\subsection{Evaluation Metrics}
To evaluate the performance of the proposed FedFT framework for Federated Learning-based Intrusion Detection Systems, we employ the following metrics:  

\begin{enumerate}
    \item \textbf{Accuracy:} The proportion of correctly classified instances (both positive and negative) relative to the total number of instances in the dataset. 

    \item \textbf{Training Memory Consumption:} The memory required to store model parameters, activations, and gradients during training. Lower memory usage is crucial for resource-constrained edge devices.

    \item \textbf{Execution Time:} The total time required to complete the training process, reflecting the computational efficiency of the framework, particularly under real-time constraints.

    \item \textbf{Scalability:} The ability of the framework to maintain performance and efficiency as the number of participating devices or the dataset size increases.
\end{enumerate}

\section{Results and Discussion}
This section presents the performance evaluation of the proposed FedFT framework, including a comparison with the baseline approaches (Table \ref{tab:perf_results1}) and an analysis of scalability (Table \ref{tab:scalability}). The results are discussed in the context of accuracy, memory usage, execution time, and scalability, highlighting trade-offs and key contributions.

\subsection{Performance Comparison}
Table \ref{tab:perf_results1} compares the FedFT framework (FedFT-1 and FedFT-3) with a traditional centralized approach and a state-of-the-art FL-IDS framework \cite{bhavsar2024fl}. FedFT-3 achieves a competitive accuracy of 99.2\%, comparable to the centralized framework (99.8\%) while reducing memory usage by 20\% and execution time by 5\%. FedFT-1, with fewer fine-tuned layers, offers significant resource savings, reducing memory usage by 42\% and execution time by 75\%, at the cost of a 5\% drop in accuracy.

\begin{table}[h!]
\centering
\caption{Performance metric comparison of the centralized framework, FL-IDS \cite{bhavsar2024fl}, and the proposed FedFT framework. FedFT-1 and FedFT-3 represent configurations with one(the last FC layer) and three fine-tuned FC layers, respectively, in the MLP module.}
\begin{tabular}{@{}lcccc@{}}
\toprule

\textbf{Framework}   & \textbf{Accuracy} &  \textbf{Loss} & \textbf{Avg. Memory } & \textbf{Avg. Execution} \\
& & &\textbf{Usage, MB} & \textbf{ Time, s}\\ \midrule
Centralized         & \textbf{99.8} &0.05   & 8.1  & 1339   \\  
FL-IDS \cite{bhavsar2024fl}          & 97.6   & 0.08  & 8.5 &1319    \\
\textbf{FedFT-3}        & 99.2   & \textbf{0.02}  & 6.6 & 1253  \\ 

\textbf{FedFT-1}         & 94.3   & 0.15   & \textbf{4.3}  & \textbf{564}  \\
\bottomrule
\end{tabular}
\label{tab:perf_results1}
\end{table}

These results demonstrate the trade-offs between resource efficiency and accuracy. Although FedFT-3 balances high accuracy and reduced resource consumption, FedFT-1 prioritizes efficiency, making it more suitable for resource-constrained environments. The ability to fine-tune specific layers provides flexibility, enabling tailored solutions based on application requirements.

\subsection{Scalability Analysis}
Scalability results are shown in Table \ref{tab:scalability}, using FedFT-1 as the default configuration. The framework maintains robust performance with minimal accuracy degradation ($<1\%$) as the number of clients increases. Memory usage remains stable, reflecting efficient resource distribution across clients. However, training execution time increases slightly due to additional communication overhead associated with more clients.

\begin{table}[h!]
\centering
\caption{Scalability results of the FedFT framework with varying client counts (Raspberry Pi 4s and virtual Flower client instances), showing accuracy, loss, memory usage, and training time.}
\begin{tabular}{ccccc}
\toprule

\textbf{No. of}   & \textbf{Accuracy} &  \textbf{Loss} & \textbf{Avg. Memory } & \textbf{Avg. Training} \\
Clients& & &\textbf{Usage, MB} & \textbf{Execution Time, s}\\ \midrule
4         & 94.8 &0.13   & 5.5  & 478   \\  
6          & 94.7   & 0.15  & 6 &472    \\
8          & 93.9   & 0.15  & 5.3 & 575 \\ 
\bottomrule
\end{tabular}
\label{tab:scalability}
\end{table}

These findings highlight the scalability of the FedFT framework, making it suitable for large-scale CAV networks. The ability to maintain efficient resource usage and consistent performance across varying client counts demonstrates the framework's adaptability to real-world deployment scenarios.

\subsection{Key Insights}
The proposed FedFT framework demonstrates competitive accuracy of up to 99.2\% while achieving significant reductions in memory usage (up to 42\%) and execution time (up to 75\%). It offers flexibility through configurable fine-tuning, enabling a balance between accuracy and resource consumption to meet specific application needs. Additionally, the framework exhibits strong scalability, maintaining performance with minimal degradation as the number of clients increases. These results validate the FedFT framework’s contributions, addressing the critical challenges of resource efficiency and scalability in Federated Learning-based Intrusion Detection Systems for connected and autonomous vehicle ecosystems.

\section{Conclusion}
In this work, we proposed a Federated Learning-based intrusion detection framework to address the dual challenges of resource constraints and scalability in connected and autonomous vehicle ecosystems and transportation IoT. Extensive evaluation demonstrated the framework’s superior performance over existing Federated Learning methods, achieving reduced memory consumption and faster training times, making it highly effective for resource-constrained environments. Scalability analysis confirmed its adaptability to increasing numbers of clients with minimal performance degradation. Although a slight decrease in accuracy was observed, this can be addressed through improved model selection. These findings highlight the framework’s practicality and robustness, contributing to advancements in IoT security and Federated Learning. Future work will explore the impact of data heterogeneity on resource usage, further refining the framework for real-world deployment.

\section{Acknowledgement}
The authors would like to thank the support from the North Carolina Department of Transportation (NCDOT) under the award number TCE2020-03. The contents do not necessarily reflect the official views or policies of NCDOT. This paper does not constitute a standard, specification, or regulation.

\bibliographystyle{IEEEtran}
\bibliography{sample}

\end{document}